\documentclass[11pt,journal,compsoc]{IEEEtran}
\usepackage{color}
\usepackage{graphicx}
\usepackage[justification=centering]{caption}

\twocolumn

\ifCLASSOPTIONcompsoc
  \usepackage[nocompress]{cite}
\else
  \usepackage{cite}
\fi

\usepackage{graphicx}
\usepackage{amsmath}
\usepackage{color}
\usepackage{amsmath, amsthm, amssymb}
\usepackage{etoolbox}
\usepackage{times}
\usepackage{epsfig}
\usepackage{float}
\usepackage{tikz}
\usepackage{comment}
\usepackage{wrapfig}
\usepackage{amsbsy} 
\usepackage{paralist}
\usepackage{multirow}
\usepackage{tabularx}
\usepackage{subcaption}
\usepackage[font=scriptsize]{caption}
\usepackage{adjustbox}

\usepackage{cases}

\DeclareMathOperator*{\argmin}{arg\,min}

\newtheorem{prop}{Proposition}

\renewcommand\thefigure{\arabic{figure}}
\newcommand{\insertfig}{
    \centering
    \includegraphics[width=0.9\textwidth]{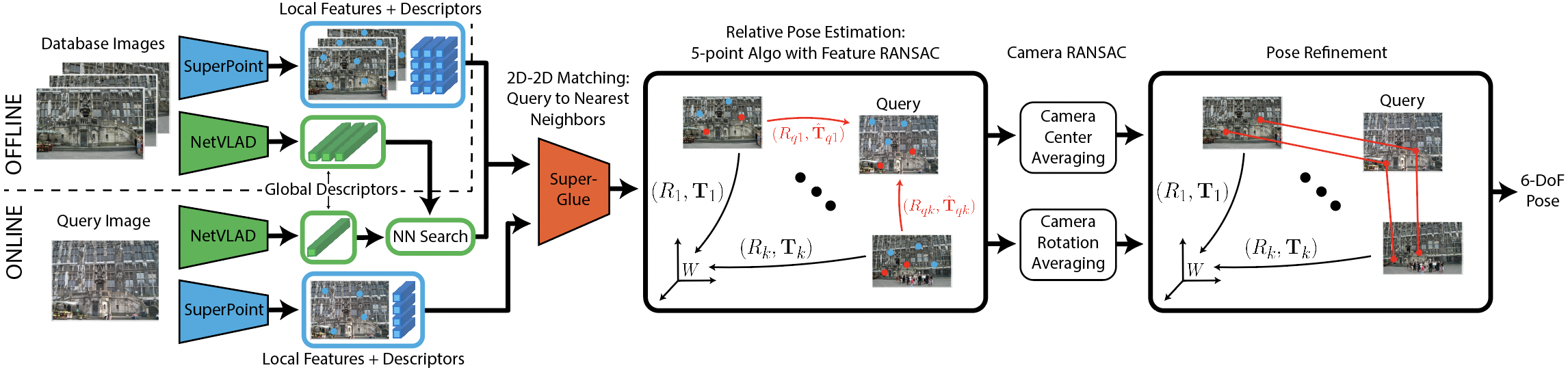}
    \captionof{figure}{An overview of our approach: the offline stage computes the NetVLAD ~\cite{arandjelovic2016netvlad} descriptors and local SuperPoint ~\cite{superpoint} features/descriptors for all database images. The online stage first computes the NetVLAD descriptors and SuperPoint features/descriptors for the query. The query NetVLAD descriptors are then used to retrieve the $K=150$ most similar images, called anchor images, by a nearest neighbor search in the descriptor space. SuperGlue ~\cite{superglue} is used to detect 2D-2D feature matches between the query and each of the $K$ anchor images. The 5-point algorithm ~\cite{5point} is used to compute relative pose between the query and each anchor image. A novel approach is used to find the optimal camera center and orientation, which are then used to find multiview feature correspondences. These multiview correspondences are the basis of a latent 3D reconstruction for highly accurate pose estimation.}
    \label{fig:main}
}

\ifCLASSINFOpdf
\else
\fi
\usepackage{graphicx} 

\makeatletter
\apptocmd{\@maketitle}{\centering\insertfig}{}{}
\makeatother

\begin{document}
%
\title{Multiview Image-Based Localization}
%
%
%
%

\author{Cameron Fiore, Hongyi Fan, Benjamin Kimia}

\IEEEtitleabstractindextext{%
\begin{abstract}
The image retrieval (IR) approach to image localization has distinct advantages to the 3D and the deep learning (DNN) approaches: it is seen-agnostic, simpler to implement and use, has no privacy issues, and is computationally efficient. The main drawback of this approach is relatively poor localization in both position and orientation of the query camera when compared to the competing approaches. This paper represents a hybrid approach that stores only image features in the database like some IR methods, but relies on a latent 3D reconstruction, like 3D methods but without retaining a 3D scene reconstruction. The approach is based on two ideas: {\em (i)} a novel proposal where query camera center estimation relies only on relative translation estimates but not relative rotation estimates through a decoupling of the two, and {\em (ii)} a shift from computing optimal pose from estimated relative pose to computing optimal pose from multiview correspondences, thus cutting out the ``middle-man''. Our approach shows improved performance on the 7-Scenes and Cambridge Landmarks datasets while also improving on timing and memory footprint as compared to state-of-the-art.
\end{abstract}
}

\IEEEdisplaynontitleabstractindextext

%
\IEEEpeerreviewmaketitle
\maketitle
\section{Introduction}
Visual localization is a highly-precise, reliable alternative~\cite{wang2021pose,choi2011localization,bejuri2015emergency} to GPS localization and important to technologies such as self-driving cars, autonomous drones, and augmented reality devices~\cite{walch2017image,zhou2020da4ad}. Visual localization is based on a database representation of the environment, {\em e.g.}, 3D point clouds~\cite{sattler2011fast,sattler2012image}, images with geo-tags~\cite{kim2015predicting,zhuang2013effective}, {\em etc}, resulting in centimeter-level precision in location and fraction-of-degree precision in orientation in both indoor and outdoor scenes~\cite{distill,DSAC,taira2018inloc,2Learn}. There are generally three canonical approaches to visual localization: {\em (i)} the image retrieval (IR) approach,   {\em (ii)} the 3D approach, and {\em (iii)} the deep learning approach. Each is described below in turn. 

\noindent {\bf The Image Retrieval Approach:} The scene is represented by a set of geo-tagged images. The top-K most similar images to the query image are retrieved. This is typically based on some global descriptors, such as NetVLAD~\cite{arandjelovic2016netvlad}, using nearest neighbor search. In the extreme cases, the pose of the most similar image is used as an estimate of the query pose. In general, this is not accurate and the relative pose of the query images with respect to multiple anchor images is used to estimate the absolute pose of the query image~\cite{sattler2018benchmarking,naseer2017semantics,2Learn}. This process is generally based on using local features with feature matching in a RANSAC scheme to acquire relative pose between the query image and anchor images followed by some form of translation/rotation averaging to find the optimal query pose. This approach is easily amenable to incremental addition of images and can span multiple scenes without additional effort, a clear advantage over scene-specific approaches. Despite its ease of implementation and use, this approach has had the distinct disadvantage of poor localization accuracy. 

\noindent \textbf{The 3D Approach:} In this approach the scene is reconstructed from image features using structure from Motion~(SfM) and represented as a 3D point cloud, with each point endowed with feature descriptors ~\cite{sattler2017large, schoenberger2016sfm,wu2011visualsfm,Im2Im,RetAndLoc}. Given a query image, its 2D features are matched to the 3D points and the query absolute pose is estimated. The accuracy of this approach is critically dependent on {\em (i)} the quality of the 3D point cloud, {\em i.e.}, whether there are a sufficient number of features, and {\em (ii)} the precision of the 3D point locations. While tremendous progress has been made in reconstructing 3D point clouds~\cite{schoenberger2016sfm,wu2011visualsfm}, this reconstruction is by no means trivial, especially for large–scale scenes when captured by non-experts~\cite{sattler2017large}: in fact \cite{charatan2021benchmarking} shows that for head-mounted cameras and indoor scenes, the state-of-the-art 3D reconstruction methods may fail when the scanning is not carefully done. Moreover, such reconstructions require significant computational resources for large-scale scenes so that incremental additions may require recomputing the scene, a challenging constraint for applications such a AR and environment exploration~\cite{arnold2022map}, or applications with embedded systems~\cite{humenberger2022investigating}. Finally, 3D point cloud representations face privacy issues since the original image can be largely recovered by 3D points and the corresponding features~\cite{pittaluga2019revealing}. There are variants that attempt to solve this issue, such as 3D lines replacing 3D points~\cite{speciale2019privacy}, but in general these face the same challenges.

\noindent {\bf The Deep Learning Approach:} A neural net is trained to regress either the relative pose of the query with respect to retrieved images (RPR) or to regress absolute pose (APR). For RPR \cite{2Learn,distill} a single image or multiple images similar to the query are retrieved and used to train a neural net that returns the relative pose of the query with respect to the similar images. Using this relative pose, the absolute pose can be calculated with epipolar geometry. Other ``end-to-end'' neural-net-based techniques learn to regress the absolute pose of the query \cite{DSAC}. These absolute pose regressors can learn from a variety of different inputs, and in the minimal case, they only need a database of images from the environment labeled with ground-truth poses. Neural-net based models have shown competitive accuracy in recent years, but training them can require huge amounts of time and training samples, and often the trained models cannot generalize well to new scenes~\cite{kendall2015posenet, kfnet, DSAC}. More recent approaches that are scene agnostic do so at the expense of accuracy~\cite{PixLoc}.

\noindent \textbf{The proposed approach} is a hybrid of these methods (see Figure 1). Like IR methods it identifies a series of most similar anchor images and optimally combines relative pose between the query image and anchor images to estimate the absolute pose. However, instead of simply using multiple relative poses to arrive at the optimal absolute pose, the relative pose estimates are removed as "middle men" and instead, the proposed methods relies on the features directly, as in the 3D approach, but without explicitly retaining a 3D reconstruction. This hybrid approach presents two key novelties: First, it proposes a decoupling between translation and orientation estimates. Rather, the camera center averaging relies solely on translation estimates, thus decoupling two potential sources of error, and leading to more accurate localization. The second key novelty is based on the assumption that the high accuracy of the 3D methods is based on leveraging feature localization over many images. The IR methods combine multiple relative pose estimates with anchor images, and thus lose the benefit of multiple view feature localization. This paper proposes to use the accurate localization of stage 1 to identify corresponding features between the query and anchor images and use these to arrive at a refined pose, at par with 3D localization accuracy. Experimental results using the 7-scenes~\cite{7Scenes}, Cambridge Landmarks~\cite{kendall2015posenet}, Aachen~\cite{acchen1, acchen2}, and RobotCar~\cite{acchen1, robot2} show across the board improvements over all methods on the Cambridge dataset, improved localization on 4 of the 7 scenes in \cite{7Scenes}, and at par performance with the state of the art on the Aachen-day/night and RobotCar-Seasons datatsets. The timing and storage requirements when compared to HLOC+SG~\cite{hloc, superglue} shows 5\% and 10\% improvement, respectively.

\section{Formulation and Notation}

The query image is denoted as $I_q$. The pose of the query camera is computed from $K$ {\em anchor images} selected from the dataset, denoted as $I_k, k=1,2,...,K$, whose absolute pose $(R_k, \mathbf{T}_k)$ is known, where $R$ and $\mathbf{T}$ are the rotation matrix and the translation vector, respectively. This means that a 3D point expressed as $\Gamma$ in the world coordinates can be expressed as $\Gamma_k$ in the coordinates of image $I_k$, where
\begin{equation}
\label{eq:1}
    \Gamma_k = R_k\Gamma + \mathbf{T}_k.
\end{equation}
Similarly, the pose of the query image $I_q$ is $(R_q,\mathbf{T}_q)$ so that $\Gamma_q = R_q\Gamma + \mathbf{T}_q$. Let $\mathbf{c}$ denote the center of the camera of an image $I$ in world coordinates, where $\mathbf{c} = -R^T \mathbf{T}$. Thus, $(\mathbf{c},R)$ is an alternate representation of pose $(R,\mathbf{T})$. The unknown pose $(R_q,\mathbf{T}_q)$ is generally what we would like to recover. This is done by finding the relative pose of the query image $I_q$ with respect to each anchor image $I_k$, $(R_{qk}, \mathbf{T}_{qk})$, which in combination with $(R_k, \mathbf{T}_k)$ provides one estimate for $(R_q, \mathbf{T}_q)$, {\em i.e.},
\begin{equation}
\begin{aligned}
    \Gamma_q &= R_{qk}\Gamma_k + \mathbf{T}_{qk} \\
    &= R_{qk}(R_k\Gamma + \mathbf{T}_k) + \mathbf{T}_{qk} \\
    &= (R_{qk}R_k)\Gamma + (R_{qk}\mathbf{T}_k + \mathbf{T}_{qk}),
\label{eq:3}
\end{aligned}
\end{equation}
so that estimate is \footnote{Note that $\Gamma_k = R_{qk}^T (\Gamma_q - \mathbf{T}_{qk}) = R_{qk}^T \Gamma_q - R_{qk}^T \mathbf{T}_{qk}$, so that $R_{kq} = R_{qk}^T$ and $\mathbf{T}_{kq} = - R_{qk}^T \mathbf{T}_{qk}$.}
\small \begin{equation}
\label{eq:forward5}
    R_q = R_{qk}R_k, \qquad \mathbf{T}_q = R_{qk}\mathbf{T}_k + \mathbf{T}_{qk}.
\end{equation} \normalsize
Unfortunately, the relative pose $(R_{qk}, T_{qk})$ computation faces metric ambiguity: denoting $T_{qk} = \lambda_{qk} \hat{T}_{qk}$ where $\hat{T}_{qk} = \frac{T_{qk}}{|T_{qk}|}$ and $\lambda_{qk} = |T_{qk}|$, $\lambda_{qk}$ is not computable and a relative pose algorithm \cite{5point} can only give $(R_{qk}, \hat{T}_{qk})$. The use of two or more cameras instead of one can settle the scale ambiguity.

\begin{figure}[h]
    \centering
    \includegraphics[width=\linewidth]{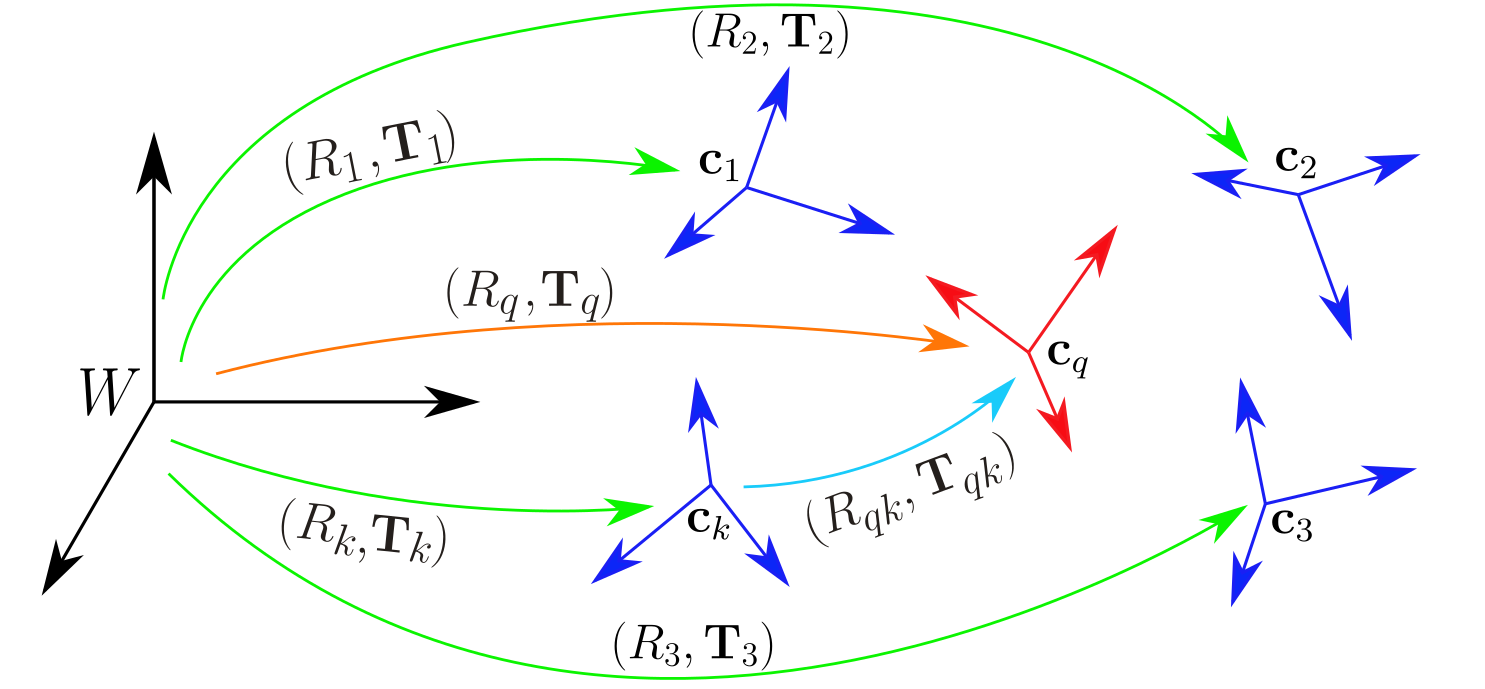}    
    \caption{The localization of the image query coordinates (red), namely, $(R_q,\mathbf{T}_q)$, with respect to the world coordinates (black) relies on the pre-localized coordinates of neighboring anchor images (blue), namely, $(R_k,\mathbf{T}_k)$, and the relative pose of the query with respect to its neighboring images, $(R_{qk},\mathbf{T}_{qk})$.}
    \label{fig:coordinate}
\end{figure}

\section{Decoupling Orientation and Translation Estimation} \label{sec:pose}

This section presents an approach to decoupling orientation and localization in arriving at an optimal pose to prevent peneatration of error between the two estimation processes.

The traditional approach to estimating a query camera pose when multiple relative pose estimates are available is to optimize the algebraic error in the translation vector $\mathbf{T}_q$ \cite{govindu}. Specifically each anchor image leads to an estimate for $\mathbf{T}_q$ which can be obtained from Equation \ref{eq:forward5}.
\begin{equation}
\mathbf{T}_q = R_{qk} \mathbf{T}_k + \lambda_{qk} \hat{T}_{qk} = R^T_{kq} \mathbf{T}_k - \lambda_{qk} R^T_{kq} \hat{\mathbf{T}}_{kq}.
\end{equation}
The free parameter $\lambda_{qk}$ can be isolated by rewriting this as $\mathbf{T}_k - R_{kq} \mathbf{T}_q = \lambda_{qk} \hat{\mathbf{T}}_{kq}$
and eliminating $\lambda_{qk}$ by taking the cross-product with $\hat{\mathbf{T}}_{kq}$,
\begin{equation} 
\label{eq:cross_prod}
    \left[ \hat{\mathbf{T}}_{kq}\right]_{\times} 
    \left(\mathbf{T}_k - R_{kq} \mathbf{T}_q\right) = 0, \qquad k = 1,2,...,K.
\end{equation}
Note that $\hat{\mathbf{T}}_{kq}$ and $R_{kq}$ are known from relative pose estimation, and $\mathbf{T}_k$ is known from the absolute pose of anchor images (in Govindu \cite{govindu}, which considers a more general problem, $\mathbf{T}_k$ is not known but the approach is the same). Equation \ref{eq:cross_prod} provides a constraint on $\mathbf{T}_q$ for each anchor image, but cannot solve for it. With additional cameras, the estimation process becomes over-constrained and~\cite{govindu} opts for a least squares solution that is completed in an iterative manner. The critical observation is that the estimation of the optimal $\mathbf{T_q}$ involves both the relative translation $\mathbf{\hat{T}_{kq}}$ and relative orientation $R_{kq}$. 

Similarly, the query orientation $R_q$ can be estimated \cite{govindu} by rewriting Equation \ref{eq:forward5} as $ R_{qk}^T R_q = R_k$ which can be converted to quaternion form,
\begin{equation} \label{quat_form}
    Q_{kq} \hat{\mathbf{q}}_q = \hat{\mathbf{q}}_k,
\end{equation}
where $\hat{\mathbf{q}}_q$ is the unit quaternion representation of $R_q$, $\hat{\mathbf{q}}_k$ is the unit quaternion representation of $R_k$, and $Q_{kq}$ represents the linear transformation from $\hat{\mathbf{q}}_q$ to $\hat{\mathbf{q}}_k$ derived from the unit quaternion representation of $R_{qk}^T$, namely 
\begin{equation}
\hat{\mathbf{q}}_{kq} = \begin{bmatrix} q_0 \\ q_1 \\ q_2 \\ q_3 \end{bmatrix} \quad
Q_{kq} = \begin{bmatrix}
    q_0 & -q_1 & -q_2 & -q_3 \\
    q_1 & q_0 & -q_3 & q_2 \\
    q_2 & q_3 & q_0 & -q_1 \\
    q_3 & -q_2 & q_1 & q_0 
\end{bmatrix}.
\end{equation}
With $K$ anchor images, Equation \ref{quat_form} gives $4K$ equations in $4K$ unknowns. The least squares solution ~\cite{govindu}, gives $\hat{\mathbf{q}}_q$ which is then converted to $R_q$. Observe that Govindu's orientation averaging estimate does not require relative pose translation $\hat{\mathbf{T}}_{qk}$ but only relative pose orientation $R_{qk}$, in contrast to Govindu's translation averaging in Equation ~\ref{eq:cross_prod} which requires both.

Before presenting the key idea of this section, we observe that the approach proposed by Markley et al. ~\cite{nasa} provides a more accurate rotation averaging by finding:
\begin{equation} 
\label{nasa_min}
    \hat{\mathbf{q}}_q = \argmin_{\hat{\mathbf{q}} \in \mathbb{S}^3} \frac{1}{K} \sum\limits_{k=1}^K ||R(\hat{\mathbf{q}}) - R_{qk}R_k) ||^2_F,
\end{equation}
where $R(\hat{\mathbf{q}})$ represents the rotation matrix corresponding to unit quaternion $\hat{\mathbf{q}}$, $||\cdot||^2_F$ is the square of the Frobenius norm, and $\mathbb{S}^3$ denotes the unit 3-sphere.

\begin{figure}[h] 
\centering
    (a)
    \includegraphics[width=.4\linewidth]{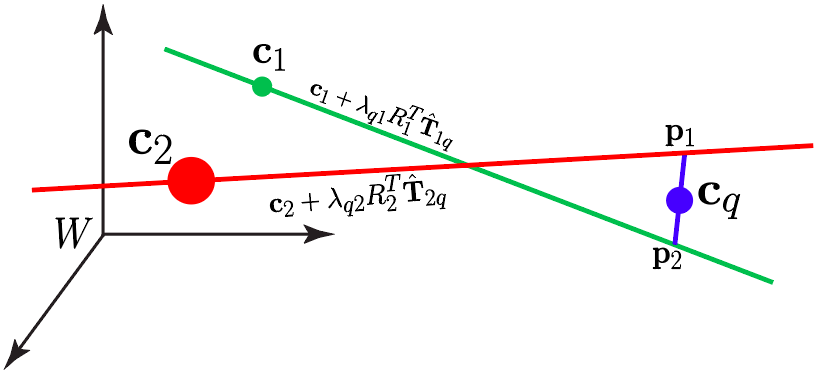}
    (b)
    \includegraphics[width=.4\linewidth]{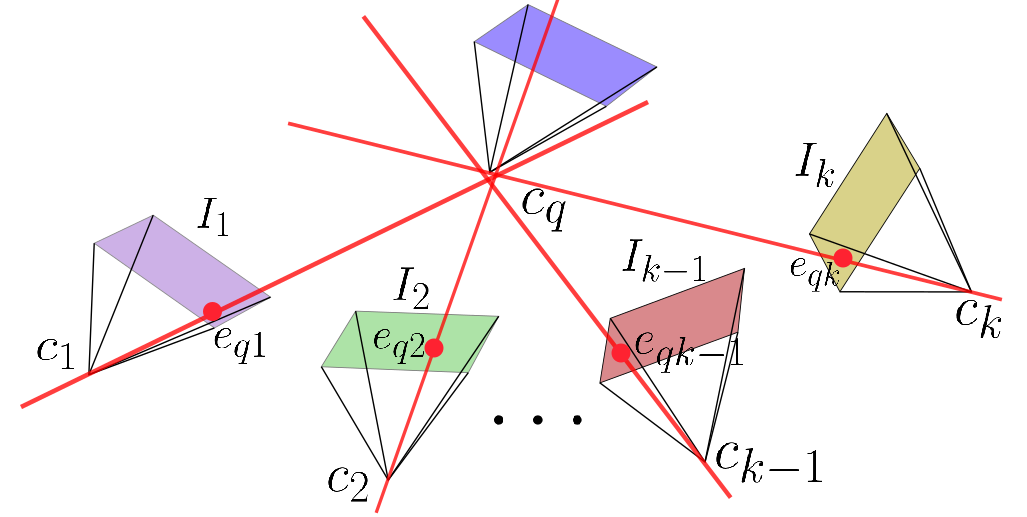}
    \caption{(a) The camera centers of $I_1$ and $I_2$, depicted above as $\mathbf{c}_1$ and $\mathbf{c}_2$, and the corresponding rays extending outward from them. These rays should intersect at the query camera center but do not in practice. The optimal query camera center can then be taken as the midpoint of the shortest line $\mathbf{p}_1 \mathbf{p}_2$ between these rays as done in triangulation. (b) When using $K$ cameras, each anchor image $I_k$ constrains the location of $\mathbf{c}_q$ to be on the line from $c_k$ to the epipole of $I_q$ on $I_k$, $e_{qk}$. The optimal $\mathbf{c}_q$ minimizes the distances to these loci. Note that this process is completely independent of the orientation of the query camera $R_q$.}
\label{fig:rays}
\end{figure}

\noindent \textbf {A Novel Geometric Approach:} 
The metric ambiguity of the relative pose between the query image $I_q$ and the anchor image $I_k$, {\em e.g.}, as computed by the Nister's five-point algorithm~\cite{5point}, \footnote{This method first compute the essential matrix $E_{qk} =[\mathbf{T}_{qk}]_\times R_{qk}$, from which four possible poses are obtained: $(R_{qk}, \hat{\mathbf{T}}_{qk})$, $(R_{qk}, -\hat{\mathbf{T}}_{qk})$, $(\Bar{R}_{qk}, \hat{\mathbf{T}}_{qk})$, $(\Bar{R}_{qk}, -\hat{\mathbf{T}}_{qk})$, where $\Bar{R}_{qk}$ is obtained by rotating the coordinates $180^\circ$ around the baseline. These choices correspond to the four possible combinations of positive or negative depths for the two cameras. The correct pose requires positive depth on both cameras which gives a unique relative pose.} and the degree of freedom in selecting $\lambda_{qk}$, implies that the query camera center lies on a one-parameter locus,
\begin{equation} 
\label{eq:5}
\begin{aligned} 
    \mathbf{c}_q (\lambda_{qk}) &= -R_q^T \mathbf{T}_q \\
    &= - (R_k^T R_{qk}^T) (R_{qk} \mathbf{T}_k + \lambda_{qk} \hat{\mathbf{T}}_{qk}) \\
    &= - R_k^T \mathbf{T}_k - \lambda_{qk} R_k^T R_{qk}^T \hat{\mathbf{T}}_{qk} \\
    &= \mathbf{c}_k + \lambda_{qk} R_k^T \hat{\mathbf{T}}_{kq},
\end{aligned}
\end{equation}
where $\mathbf{c}_k = -R_k^T \mathbf{T}_k$ and $\mathbf{T}_{kq} = - R_{qk}^T \mathbf{T}_{qk}$. Observe that this locus is a 3D line passing through $\mathbf{c}_k$ along the unit direction $R_k^T \hat{\mathbf{T}}_{kq}$. It is also important to note that this line also goes through the epipole of the query camera on the anchor image $e_{qk}$, namely, the projection of the query camera center $\mathbf{c}_q$ onto the anchor image $I_k$, Figure \ref{fig:rays}(a). When $K$ anchor images are available, each anchor image insists that the center be on this 3D line, so that an optimal center minimizes the displacements from these $K$ lines. 
Consider first the case, $K=2$, to illustrate this. From Equation~\ref{eq:5} two estimates for $\mathbf{c}_q$ emerge, namely,
\begin{equation}
\begin{aligned}
    &\mathbf{c}_q = \mathbf{c}_1 + \lambda_{q1} R_1^T \hat{\mathbf{T}}_{1q}, \\
    &\mathbf{c}_q = \mathbf{c}_2 + \lambda_{q2} R_2^T \hat{\mathbf{T}}_{2q}.
\end{aligned}
\end{equation}
Each anchor image postulates a 3D line as the locus of the query camera center, but the two loci generally do not intersect, Figure~\ref{fig:rays}(a). From a geometric perspective, the best estimate can be argued to be the midpoint of the shortest line connecting the loci, and this is the that is followed by~\cite{2Learn} who use two cameras to estimate camera pose. For three cameras, the optimal center lies on a triangle whose vertices are on the three rays. The situation quickly becomes more complex for more cameras, motivating an algebraic approach to minimize the distance between $\mathbf{c}_q$ and the rays it is supposed to lie on, Figure~\ref{fig:rays}(b). 

\begin{prop} \label{prop:prop_c_q}
The query camera center $\mathbf{c}_q$ which minimizes the sum of squared distances between $\mathbf{c}_q$ and the rays defined by $\mathbf{c}_k + \lambda_{qk} R_k^T \hat{\mathbf{T}}_{kq}, k=1,2,...,K$ is computed from
\small
\begin{equation}
\label{eq: prop_1a}
    \left(\sum_{k=1}^{K} \left(I - R_k^T \hat{\mathbf{T}}_{kq} \hat{\mathbf{T}}_{kq}^T R_k\right) \right)\mathbf{c}_q = \sum_{k=1}^{K} (I - R_k^T \hat{\mathbf{T}}_{kq} \hat{\mathbf{T}}_{kq}^T R_k) \mathbf{c}_k. 
\end{equation}
\normalsize
\end{prop}
\noindent The proof to Proposition \ref{prop:prop_c_q} is given in the supplementary material. This proposition gives a closed-form solution of the query camera center, computed from the pose of anchor images and it is the core to our algorithm. The critical observation is that the optimal camera center computation only involves translation estimation relative to anchor images and not relative orientation estimates.

\noindent \textbf{The Proposed Algorithm:} 

\noindent {\em (i)} \textbf{ Retrieve Top $\mathbf{K}$ Anchor Images:} The $K$ most similar images to the query image are found using NetVLAD descriptors~\cite{arandjelovic2016netvlad}.

\noindent {\em (ii)} \textbf{RANSAC:} Given that the $K$ anchor images can occasionally include outliers, a pair of images (the minimum number of anchor images that can determine query camera pose) is used in a RANSAC loop to form an approximate query camera pose hypothesis and enumerate the number of other poses consistent with this hypothesis. That set of two images which have the most number of inliers together with these $\hat{K}$ inliers then form a basis of anchor cameras from which the optimal query camera pose is computed.

\noindent {\em (iii)} \textbf{Optimal Camera Center Estimation:} Proposition~\ref{prop:prop_c_q} is used to find the optimal camera center. Observe that relative orientation $R_{kq}$ is not involved in this process.

\noindent {\em (iv)} \textbf{Optimal Camera Orientation Estimation:} Using the closed form solution proposed in ~\cite{nasa} to find $\hat{\mathbf{q}}_q$, we use $R(\hat{\mathbf{q}}_q) = R_q$ as our estimate for the absolute query rotation. Observe that $T_{qk}$ is not involved in this process.

This decoupled approach provides an accurate initial query pose estimate which identifies corresponding features which are used to directly estimate query pose, as described next.

\section{Pose Refinement via Latent 3D Points}
\label{sec:absolute}
The query pose estimation process described above is a two step procedure: {\em (i)} relative pose estimation between the query and each anchor image; {\em (ii)} optimal query pose estimation by averaging relative poses. The relative pose estimate is effectively a ``middle-man.'' The second key idea of this paper is that the integration of relative pose with multiple anchor images should be done directly based on the corresponding multiview features, which, in the first place, were the basis of relative pose estimation. This effectively cuts the middle-man out and is expected to improve localization. 

Specifically, let $\gamma_j, j=1,2,\ldots,M$ denote features in the query image, and let $\gamma_{k,j}$ be the feature corresponding to the query feature $\gamma_j$ in the $k^{th}$ view. Sometimes, for uniformity, we use the notion $\gamma_{0,j}$ to denote $\gamma_j$, effectively considering the query as the zeroth view. The feature correspondence is established when the relative pose of the query image relative to the $k^{th}$ anchor view is estimated. Note that some feature correspondences may be missing from some anchor images, but this does not change our analysis. Assume, for simplicity, that each query feature has a corresponding feature in each anchor image. The goal is to seek a query pose that is most consistent with the projection of each of the $M$ query features in each of the $N$ anchor views. Specifically, each set of corresponding features in anchor images $\gamma_{k,j}$, $k=1,2,\ldots,N$ is in correspondence with the query image feature $\gamma_j$. Collectively referred to by $\{\gamma_{k,j}$, $k=0,1,2,\ldots,N\}$, this set of image features must all arise from a single 3D point. In practice, however, the observed features $\gamma_{k,j}$ experience noise $\Delta \gamma_{k,j}$ from the true location $\hat{\gamma}_{k,j}$, where
\begin{equation} 
\label{sum_gammas}
    \gamma_{k,j} = \hat{\gamma}_{k,j} + \Delta \gamma_{k,j}, j = 1,2,...,M; k = 1,2,...,N,
\end{equation}
where the true features $\{ \hat{\gamma}_{k,j}$, $k=0,1,2,\ldots,N \}$ do arise from a single true 3D point $\hat{\Gamma}_j$, as expressed in world coordinates. Since these 3D points are unknown, they can be found by minimizing the perturbation of observed feature points from the projection of $\gamma_{k,j}$, namely, minimize
\begin{equation}
\label{demand}
    E_1 (\hat{\Gamma}_j) = \sum_{k=1}^N |\Delta \gamma_{k,j}|^2, \quad j = 1,2,...,M.
\end{equation}
This is effectively $N$-view triangulation of a set of features, which can be obtained as follows.
\begin{prop}
\label{prop:3eqs3unkwns}
The optimal $N$-view triangulation of $N$ feature points in $N$ views, $\gamma_j$, $j=1,2,\ldots,N$, based on minimizing total perturbation $E=\sum_{k=1}^{N} \left| \Delta \gamma_k \right|^2$ where $\Delta \gamma_k = \gamma_k - \hat{\gamma}_k$, is $\Gamma$ whose expression in the first view $\hat{\Gamma}_1 = \hat{\rho}_1 \hat{\gamma}_1$ can be found by minimizing 
\begin{equation}
\begin{aligned}
     &E \left(\hat{\gamma}_1,\hat{\rho}_1 \right) = \left| \gamma_1 - \hat{\gamma_1} \right|^2 \\
     &+ \sum_{k=2}^{N} \Biggl[
        \left| 
            \gamma_k \right|^2 
            - 2 \frac{\hat{\rho}_1 \gamma_k^T R_{k1} \hat{\gamma}_1 + \gamma_k^T \mathbf{T}_{k1}}{\hat{\rho}_1 e_3^T R_{k1} \hat{\gamma}_1 + e_3^T \mathbf{T}_{k1}}\\
     &\hspace{2.05cm} + \frac{\hat{\rho}_1 \left|\hat{\gamma}_1 \right|^2 + \left|\mathbf{T}_{k1} \right|^2 + 2\hat{\rho}_1 \hat{\gamma}_1^T R_{k1}^T \mathbf{T}_{k1}}{\left(\hat{\rho}_1 e_3^T R_{k1} \hat{\gamma}_1 + e_3^T \mathbf{T}_{k1} \right)^2} \Biggr].
\end{aligned}
\end{equation}
\end{prop}
\noindent The above process gives a set of 3D points from $\gamma_{k,j}$ which are expressed as $\hat{\Gamma}_j$ in world coordinates. The expression in the $k^{th}$ image coordinate is $R_k \hat{\Gamma}_j + \mathbf{T}_k$, while in the query image coordinates its expression is $R_q \hat{\Gamma}_j + \mathbf{T}_q$. The next task is to find the query pose $(R_q, \mathbf{T}_q)$ which minimizes the perturbation of query features $\gamma_{0,j} = \hat{\gamma}_{0,j} + \Delta \gamma_{0,j}$, $j=1,2,\ldots,M$ that brings the query features in register with the 3D points $\hat{\Gamma}_j$, {\em i.e.}, minimize $E_2$,
\begin{equation} 
\label{oj}
    E_2 (R_q, \mathbf{T}_q, \hat{\Gamma}_1, \hat{\Gamma}_2, ..., \hat{\Gamma}_M) = \sum_{j=1}^M |\Delta \gamma_{0,j}|^2.
\end{equation}
Thus, the overall optimization is over the query pose unknown with the 3D points $\hat{\Gamma}_j$, $j=1,2,\ldots,M$, as latent variables {\em i.e.},

\begin{subnumcases}{}
    \min_{R_q, \mathbf{T}_q} E_2 (R_q, \mathbf{T}_q, \hat{\Gamma}_1, \hat{\Gamma}_2, \ldots, \hat{\Gamma}_M) \ \ \text{subject to} \\
    \min_{\Gamma_j} E_1 (\hat{\Gamma}_j), \ j=1,2,\ldots,M.
\end{subnumcases}
Having solved for the latent 3D points $\hat{\Gamma}_j$, $j=1,2,\ldots,M$, in Proposition \ref{prop:3eqs3unkwns}, the optimal query pose can be obtained by minimizing the total discrepancy.

\begin{figure}[h]
    \centering
    \includegraphics[width=\linewidth]{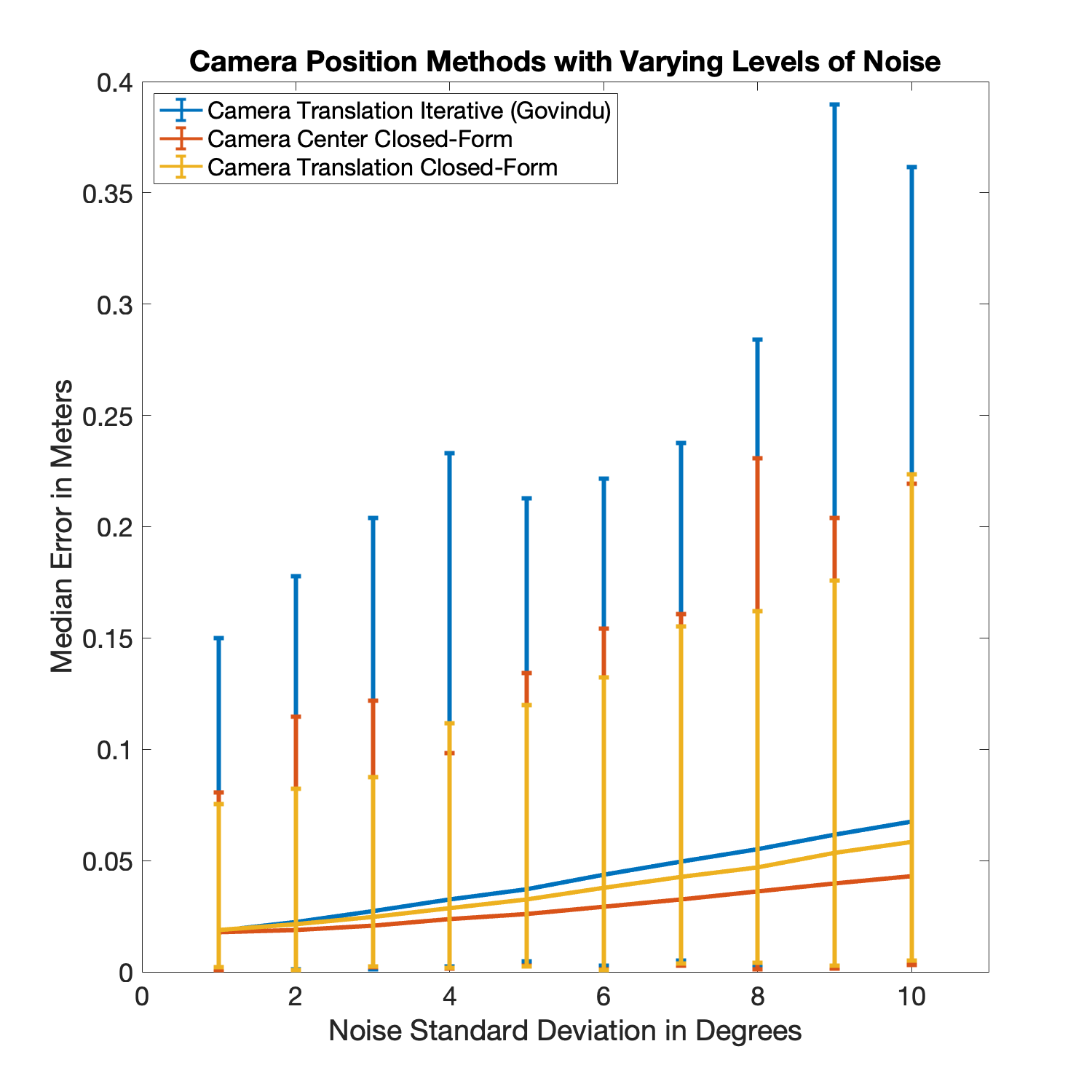}
    \caption{The localization accuracy of our method is superior to that of Govindu's iterative method~\cite{govindu} under varying degrees of simulated relative pose noise, with the disparity growing with the extent of noise.}
    \label{fig:noise}
\end{figure}

\section{Experiments}
The experiments probe the efficacy of the two key ideas presented here, namely, decoupled camera center averaging and latent 3D pose estimation. First, Figure~\ref{fig:noise} examines the localization accuracy of our approach (closed form, camera center averaging) with that of Govindu~\cite{govindu} under varying amounts of noise. In addition, through a similar analysis, closed-form translation averaging is explored and its results shown. More specifically, based on the rational that feature noise perturbs relative pose estimates between anchor and query images, which in turn impacts the localization of the query image, in a synthetic experiment each relative pose estimate $(R, \mathbf{T})$ is perturbed as $(N_R R, N_T \mathbf{T})$, where the matrices  $N_R$ and $N_T$ represents zero-mean Gaussian noise with varying standard deviation. Figure \ref{fig:noise} shows that our approach is more accurate under noise than Govindu \cite{govindu} and the disparity grows with the extent of noise.

In addition, a holistic localization experiment on a subset of the 7-Scenes and Cambridge datasets reveal that replacing Govindu's translation averaging with our camera center averaging method improves localization, as shown in Table~\ref{tab:transl-camera-averaging}.

\begin{table}
\centering
    \centering
        \begin{tabular}{|c|c|c|c|c|}
        \hline
        & Chess & Stairs & King's & St.Mary's \\
        \hline
        Translation Averaging & 4 & 11 & 20 & 17 \\
        \hline
        Center Averaging & 3 & 10 & 18 & 13 \\
        \hline
        \end{tabular}
    \caption{The localization error (cm) for two sequences each from the 7-Scenes and Cambridge datasets reveals that camera center averaging improves localization.}
    \label{tab:transl-camera-averaging}
\end{table}

The experiments probing the second key idea are evaluated on four datasets. First, the 7-Scenes Dataset \cite{7Scenes} which contains 26,000 reference images depicting seven indoor scenes. Second, the Cambridge Landmarks dataset \cite{kendall2015posenet} contains 5,365 reference images depicting 5 scenes from the city of Cambridge, UK on a much larger metric scale. Third, the Aachen Day-Night dataset \cite{acchen1, acchen2} contains 4,328 reference images depicting the old inner city of Aachen Germany, all taken with mobile phone cameras, and it offers both day and night queries. Finally, the RobotCar Seasons dataset \cite{acchen1, robot2} contains 26,121 reference images depicting the city of Oxford, UK and captured by cameras mounted on a moving car.

\begin{table*}[h]
\begin{tiny}
    \centering
    \begin{adjustbox}{max width=\textwidth}
    \begin{tabularx}{500pt}{c|l|
    >{\centering\arraybackslash}X|
    >{\centering\arraybackslash}X|
    >{\centering\arraybackslash}X|
    >{\centering\arraybackslash}X|
    >{\centering\arraybackslash}X|
    >{\centering\arraybackslash}X|
    >{\centering\arraybackslash}X|
    >{\centering\arraybackslash}X||
    >{\centering\arraybackslash}X|
    >{\centering\arraybackslash}X|
    >{\centering\arraybackslash}X|
    >{\centering\arraybackslash}X|
    >{\centering\arraybackslash}X|
    >{\centering\arraybackslash}X|}
    
    \cline{3-16}
    \multicolumn{2}{c|}{} & \multicolumn{8}{c||}{7 Scenes} & \multicolumn{6}{c|}{Cambridge Landmarks}\\
    \cline{3-16}
     
    \multicolumn{1}{c}{} & \multicolumn{1}{c|}{Pipeline} & Chess & Fire & Heads & Office & Pumpkin & Kitchen & Stairs & Average & Court & King's & Hospital & Shop & St.Mary's & Average\\
     
    \cline{2-16}
    \multirow{4}{.01em}{\rotatebox{90}{IR}}
    & DenseVLAD~\cite{torii201524} & 
    12.5/0.21 & 
    13.8/0.33 & 
    14.9/0.15 & 
    11.2/0.28 &
    11.3/0.31 & 
    12.3/0.30 & 
    15.8/0.25 &
    13.1/0.26 &
    - &
    5.72/2.80 & 
    7.13/4.01 & 
    7.61/1.11 &
    8.00/2.31 &
    7.12/2.56 \\
    & DenseVLAD+Inter.~\cite{inter} & 
    10.0/0.18 & 
    12.4/0.33 & 
    14.3/0.14 & 
    10.1/0.25 &
    11.6/0.43 & 
    13.4/0.42 & 
    14.2/0.34 &
    12.3/0.30 &
    - &
    4.45/1.48 & 
    4.63/2.68 & 
    4.32/0.90 &
    6.06/1.62 & 
    4.87/1.67\\
    & GeM \cite{gem}& 
    13.1/0.27 & 
    15.8/0.39 & 
    15.7/0.21 & 
    11.7/0.37 &
    11.6/0.43 & 
    13.4/0.42 & 
    14.2/0.34 &
    13.6/0.35 &
    - &
    - &
    - &
    - &
    - & 
    -\\
    & Zhou et al ~\cite{2Learn}& 
    1.48/0.04 & 
    1.62/0.05 & 
    1.80/0.04 & 
    1.59/0.06 &
    1.86/0.08 & 
    1.86/0.07 &
    3.69/0.22 &
    1.99/0.08 &
    - &
    0.71/0.48 & 
    1.24/0.88 & 
    0.55/0.17 &
    1.00/0.35 & 
    0.88/0.47\\
    
    \cline{2-16}
    \multirow{5}{0.01em}{\rotatebox{90}{3D}}
    & Active Search~\cite{active_search} & 
    1.96/0.04 & 
    1.53/\textcolor{red}{0.03} & 
    1.45/\textcolor{red}{0.02} & 
    3.61/0.09 &
    3.10/0.08 & 
    3.37/0.07 &
    2.22/\textbf{\textcolor{red}{0.03}} &
    2.46/0.05 &
    0.13/0.24 &
    0.22/0.13 & 
    0.36/0.20 &
    0.21/0.04 &
    0.25/0.08 & 
    0.23/0.14\\
    & GAM ~\cite{GAM} &
    - &
    - &
    - &
    - &
    - &
    - &
    - &
    - &
    0.08/0.11 &
    \textcolor{red}{0.10}/\textbf{\textcolor{red}{0.05}} &
    \textbf{\textcolor{red}{0.22}}/\textbf{\textcolor{red}{0.11}} &
    \textcolor{red}{0.13}/\textcolor{red}{0.03} &
    \textcolor{red}{0.11}/\textcolor{red}{0.04} & 
    \textcolor{red}{0.13}/\textcolor{red}{0.07} \\
    & InLoc~\cite{InLoc} & 
    1.05/\textcolor{red}{0.03} &
    1.07/\textcolor{red}{0.03} &
    1.16/\textcolor{red}{0.02} &
    1.05/\textcolor{red}{0.03} &
    1.55/\textcolor{red}{0.05} &
    1.31/\textcolor{red}{0.04} &
    2.47/0.09 &
    1.38/\textcolor{red}{0.04} &
    0.62/1.20 &
    0.82/0.46 &
    0.96/0.48 &
    0.50/0.11 & 
    0.63/0.18 & 
    0.71/0.49 \\
    & HLoc~\cite{hloc}& 
    0.79/\textbf{\textcolor{red}{0.02}} & 
    0.87/\textbf{\textcolor{red}{0.02}} & 
    0.92/\textcolor{red}{0.02} & 
    0.91/\textcolor{red}{0.03} &
    1.12/\textcolor{red}{0.05} & 
    1.25/\textcolor{red}{0.04} & 
    1.62/0.06 &
    1.07/\textbf{\textcolor{red}{0.03}} &
    0.21/0.38 &
    0.31/0.17 &
    0.39/0.23 &
    0.37/0.07 &
    0.29/0.10 & 
    0.31/0.19\\
    & HLoc + SG ~\cite{hloc} ~\cite{superglue} &
    0.80/\textbf{\textcolor{red}{0.02}} &
    0.77/\textbf{\textcolor{red}{0.02}} &
    \textcolor{red}{0.79}/\textbf{\textcolor{red}{0.01}} &
    \textcolor{red}{0.80}/\textcolor{red}{0.03} &
    \textbf{\textcolor{red}{1.07}}/\textbf{\textcolor{red}{0.04}} &
    \textbf{\textcolor{red}{1.13}}/\textbf{\textcolor{red}{0.03}} &
    \textbf{\textcolor{red}{1.15}}/\textcolor{red}{0.04} &
    \textbf{\textcolor{red}{0.93}}/\textbf{\textcolor{red}{0.03}} &
    \textcolor{red}{0.07}/\textcolor{red}{0.10} &
    \textcolor{red}{0.11}/\textcolor{red}{0.07} &
    0.24/\textcolor{red}{0.13} &
    0.14/\textcolor{red}{0.03} &
    0.12/\textcolor{red}{0.04} & 
    0.14/\textcolor{red}{0.07} \\
    
    \cline{2-16}
    \multirow{2}{0.01em}{\rotatebox{90}{DNN}}
    & DSM ~\cite{dsm} &
    \textcolor{red}{0.71}/\textbf{\textcolor{red}{0.02}} &
    0.85/\textbf{\textcolor{red}{0.02}} &
    0.85/\textbf{\textcolor{red}{0.01}} & 
    0.84/\textcolor{red}{0.03} &
    1.16/\textbf{\textcolor{red}{0.04}} &
    \textcolor{red}{1.17}/\textcolor{red}{0.04} &
    1.33/0.05 &
    0.99/\textbf{\textcolor{red}{0.03}} &
    0.23/0.44 &
    0.36/0.19 &
    0.39/0.24 &
    0.38/0.07 &
    0.35/0.12 & 
    0.34/0.21 \\
    & PixLoc ~\cite{PixLoc} & 
    0.80/\textbf{\textcolor{red}{0.02}} &  
    \textbf{\textcolor{red}{0.73}}/\textbf{\textcolor{red}{0.02}} &
    0.82/\textbf{\textcolor{red}{0.01}} &
    0.82/\textcolor{red}{0.03} &
    1.21/\textbf{\textcolor{red}{0.04}} &
    1.20/\textbf{\textcolor{red}{0.03}} &
    \textcolor{red}{1.30}/0.05 &
    \textcolor{red}{0.98}/\textbf{\textcolor{red}{0.03}} &
    0.14/0.30 &
    0.24/0.14 &
    0.32/0.16 & 
    0.23/0.05 & 
    0.34/0.10 & 
    0.25/0.15 \\

    \cline{2-16}
    & \textbf{Ours} & 
    \textbf{\textcolor{red}{0.68}}/\textbf{\textcolor{red}{0.02}} &
    \textcolor{red}{0.76}/\textbf{\textcolor{red}{0.02}} &
    \textbf{\textcolor{red}{0.56}}/\textbf{\textcolor{red}{0.01}} &
    \textbf{\textcolor{red}{0.71}}/\textbf{\textcolor{red}{0.02}} &
    \textcolor{red}{1.10}/\textbf{\textcolor{red}{0.04}} &
    1.18/\textcolor{red}{0.04} &
    1.99/0.07 &
    1.00/\textbf{\textcolor{red}{0.03}} &
    \textbf{\textcolor{red}{0.05}}/\textbf{\textcolor{red}{0.09}} &
    \textbf{\textcolor{red}{0.09}}/\textcolor{red}{0.07} &
    \textbf{\textcolor{red}{0.20}}/\textbf{\textcolor{red}{0.11}} &
    \textbf{\textcolor{red}{0.10}}/\textbf{\textcolor{red}{0.02}} &
    \textbf{\textcolor{red}{0.09}}/\textbf{\textcolor{red}{0.03}} & 
    \textbf{\textcolor{red}{0.11}}/\textbf{\textcolor{red}{0.06}}\\
    
    \cline{2-16}
    \end{tabularx}
    \end{adjustbox}
    
    \caption{Image localization results of our approach is compared to others on the 7-Scenes and Cambridge Landmarks datasets. Accuracy is reported as a pair of median errors in degrees/meters for 7-Scenes and Cambridge. Our method (an image retrieval-based method) and other IR pipelines are grouped, the 3D pipelines are grouped, and the Deep Neural Network (DNN) pipelines are grouped. The best results per group are in \textbf{\textcolor{red}{bold red}}, and the second best results are in \textcolor{red}{red}.}
    \label{tab:results_7}
\end{tiny}
\end{table*}

\begin{table}[h]
\begin{tiny}

    \centering
    \begin{tabularx}{250pt}{c|l|
    >{\centering\arraybackslash}X|
    >{\centering\arraybackslash}X||
    >{\centering\arraybackslash}X|
    >{\centering\arraybackslash}X|}

    \cline{3-6}
    \multicolumn{2}{c|}{} & \multicolumn{2}{c||}{Aachen Day-Night} & \multicolumn{2}{c|}{RobotCar Seasons}\\
    \cline{3-6}
    
    \multicolumn{1}{c}{} & \multicolumn{1}{c|}{Pipeline} & Day & Night & Day & Night\\
    \cline{2-6}
    
    \multirow{2}{.01em}
    {\rotatebox{90}{IR}}
    & DenseVLAD~\cite{torii201524} & 
    0.0/0.1/22.8 & 
    0.0/1.0/19.4 &
    7.6/31.2/91.2 &
    1.0/4.4/22.7 \\
    & NetVLAD ~\cite{arandjelovic2016netvlad} &
    0.0/0.2/18.9 &
    0.0/0.0/14.3 &
    6.4/26.3/90.9 &
    0.3/2.3/15.9 \\
    
    \cline{2-6}
    \multirow{5}{0.01em}{\rotatebox{90}{3D}}
    & Active Search~\cite{active_search} & 
    85.3/92.2/97.9 &
    39.8/49.0/64.3 &
    50.9/80.2/96.6 &
    6.9/15.6/31.7 \\
    & GAM ~\cite{GAM} &
    \textcolor{red}{88.0}/\textcolor{red}{94.8}/\textcolor{red}{98.5} &
    78.6/\textcolor{red}{91.8}/\textcolor{red}{99.0} &
    \textbf{\textcolor{red}{57.8}}/\textcolor{red}{81.6}/97.4 &
    12.6/35.3/\textcolor{red}{69.4} \\
    & CSL ~\cite{CSL} &
    52.3/80.0/94.3 &
    29.6/40.8/56.1 &
    45.3/73.5/90.1 &
    0.6/2.6/7.2 \\
    & D2-Net ~\cite{D2}&
    84.3/91.9/96.2 &
    75.5/87.8/95.9 &
    54.5/80.0/95.3 &
    \textcolor{red}{20.4}/\textcolor{red}{40.1}/55.0 \\
    & HLOC + SG ~\cite{hloc} ~\cite{superglue} & 
    \textbf{\textcolor{red}{89.6}}/\textbf{\textcolor{red}{95.4}}/\textbf{\textcolor{red}{98.8}} &
    \textbf{\textcolor{red}{86.7}}/\textbf{\textcolor{red}{93.9}}/\textbf{\textcolor{red}{100}} &
    \textcolor{red}{56.9}/\textbf{\textcolor{red}{81.7}}/\textcolor{red}{98.1} &
    \textbf{\textcolor{red}{33.3}}/\textbf{\textcolor{red}{65.9}}/\textbf{\textcolor{red}{88.8}} \\
    
    \cline{2-6}
    \multirow{2}{0.01em}{\rotatebox{90}{DNN}}
    & ESAC ~\cite{esac} &
    42.6/59.6/75.5 &
    6.1/10.2/18.4 &
    - &
    - \\
    & PixLoc ~\cite{PixLoc} &
    84.6/92.4/98.2 &
    69.4/87.8/95.9 &
    56.8/81.4/\textbf{\textcolor{red}{98.6}} &
    8.8/25.6/58.2 \\
    
    \cline{2-6}
 
    & \textbf{Ours} & 
    86.2/93.2/97.9 &
    \textcolor{red}{82.7}/\textcolor{red}{91.8}/\textbf{\textcolor{red}{100}} &
    51.8/75.6/91.5 &
    16.3/32.2/50.6 \\

    \cline{2-6}
    \end{tabularx}

    \caption{Image localization results of our approach is compared to others on Aachen Day-Night and RobotCar Seasons datasets. We report the percentage of query localized at three different accuracy levels (0.25m, 2°)/(0.5m, 5°)/(5m, 10°). Our method (an image retrieval-based method) and other IR pipelines are grouped, the 3D pipelines are grouped, and the Deep Neural Network (DNN) pipelines are grouped. The best results per group are in \textbf{\textcolor{red}{bold red}}, and the second best results are in \textcolor{red}{red}.}
    
    \label{tab:results_aachen_robot}
\end{tiny}
\end{table}

\begin{figure}[h]
    \centering
    (a) \includegraphics[width=.35\linewidth]{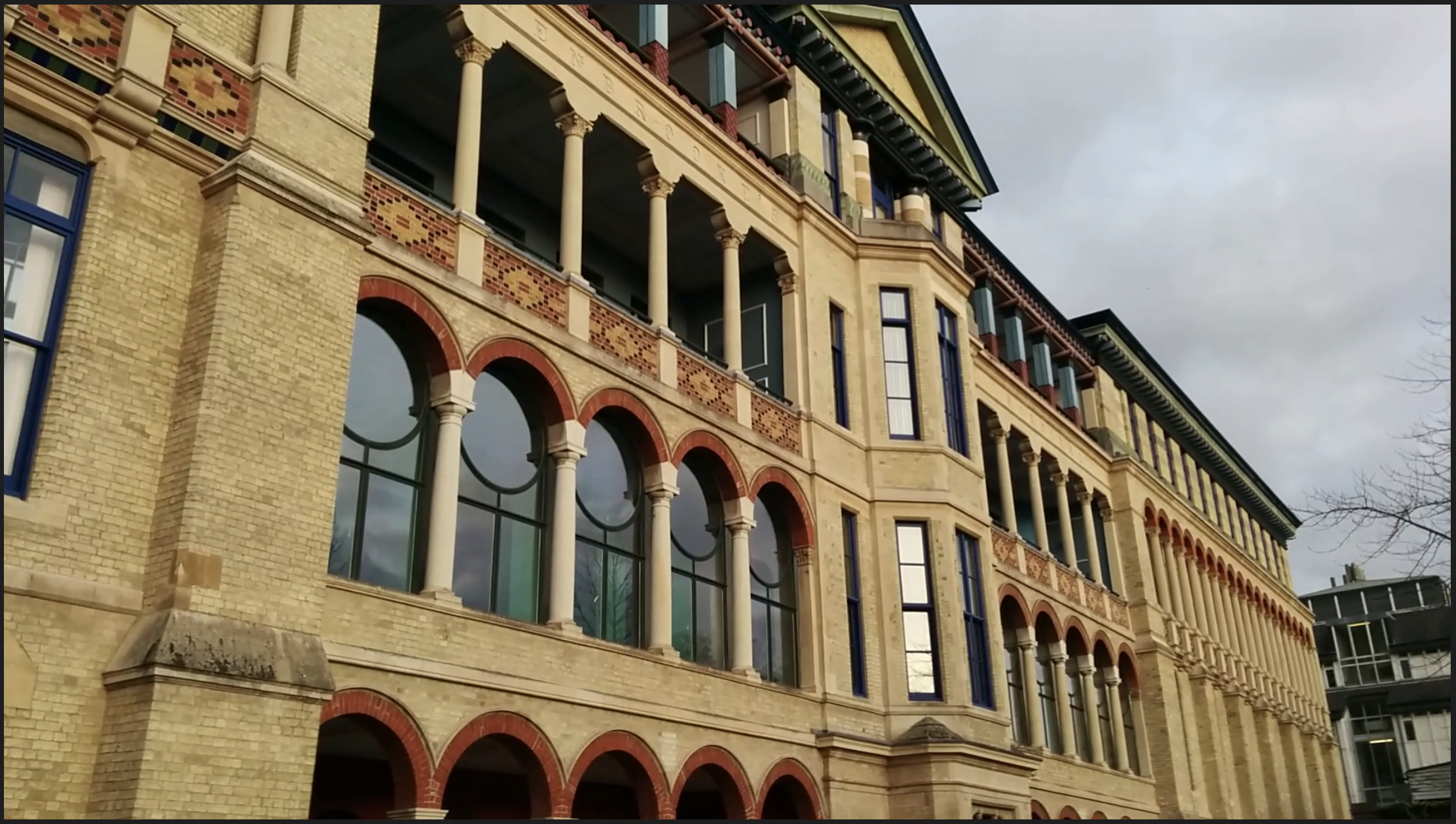}
    (b) \includegraphics[width=.5\linewidth]{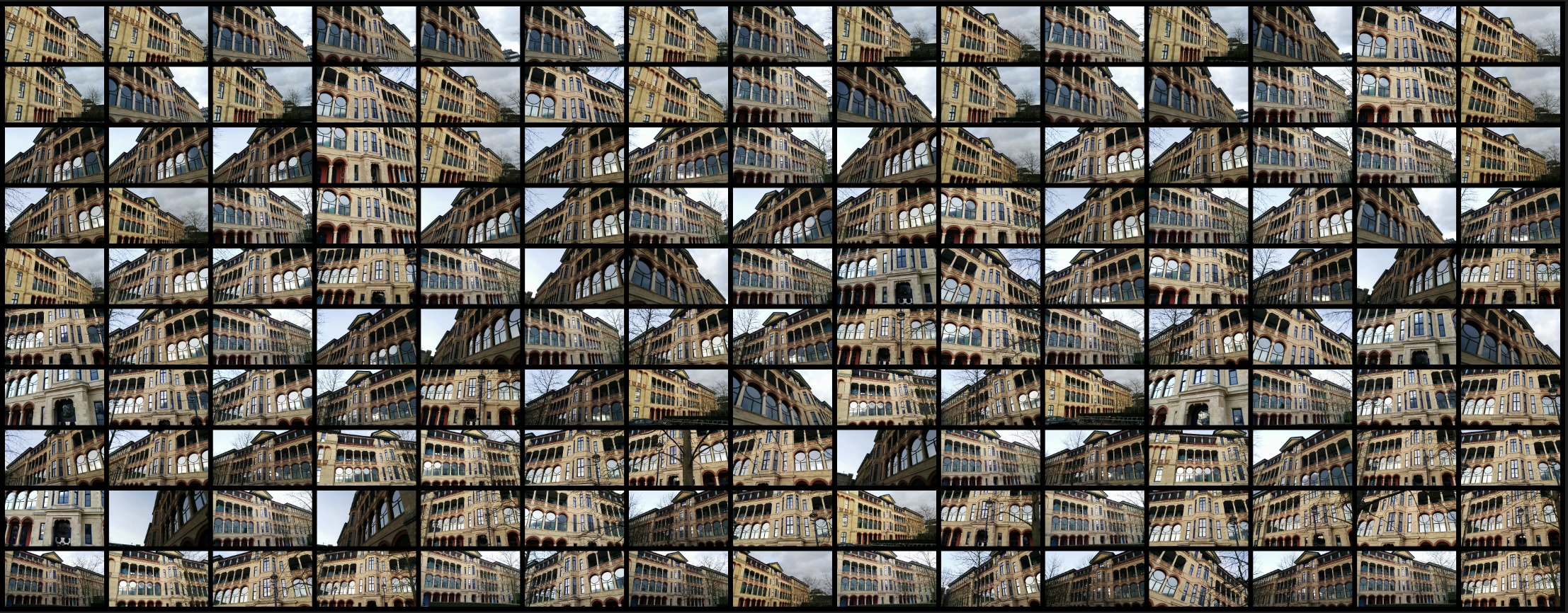}
    \caption{Example result of image retrieval step. Image (a) is the query image take from the Old Hospital scene of Cambridge Landmarks and image (b) is a $10 \times 15$ grid showing the top 150 most similar images, the vast majority of which are correct.}
    \label{fig:retrieval}
\end{figure}

\noindent \textbf{First Stage Recall:}
The first stage, image retrieval, recalls $K=150$ images from the entire dataset. Several state-of-the-art approaches are scene-specific in that the query is directed to the dataset related to a particular scene \cite{DSAC, DSAC++, kfnet}, while others are scene-agnostic in that the query is unaware of which particular scene it is related to and is expected to localize without this knowledge~\cite{esac, CSL, arandjelovic2016netvlad, PixLoc, dsm, hloc, InLoc, GAM, active_search, 2Learn, gem, inter, torii201524}. Our approach belongs to the more challenging scene-agnostic category. Figure \ref{fig:retrieval} shows that the vast majority of the recalled images are included in the vicinity of the query location and can be used to aid in localization. Figure \ref{fig:noise}(b) shows the percentage of correctly recalled images in the top 150 most similar images for the 7-Scenes, Cambridge, Aachen, and RobotCar datasets.

\noindent Observe that the distribution for 7-Scenes and Cambridge dataset shows that a vast majority of the 150 images are nearby images, while for the Aachen and RobotCar datasets this percentage is significantly fewer. This is simply due to a lack of availability of this data, and not due to poor IR performance. As we shall see in the localization results our method's localization performance is dependent on the {\em availability of a large number of images}, hence performing significantly better on the 7-Scenes and Cambridge datasets as compared to the latter two datasets for this reason. 

\begin{figure}[h]
    \vspace{0.2cm}
    \centering{
    (a) \fbox{\includegraphics[width=.4\linewidth]{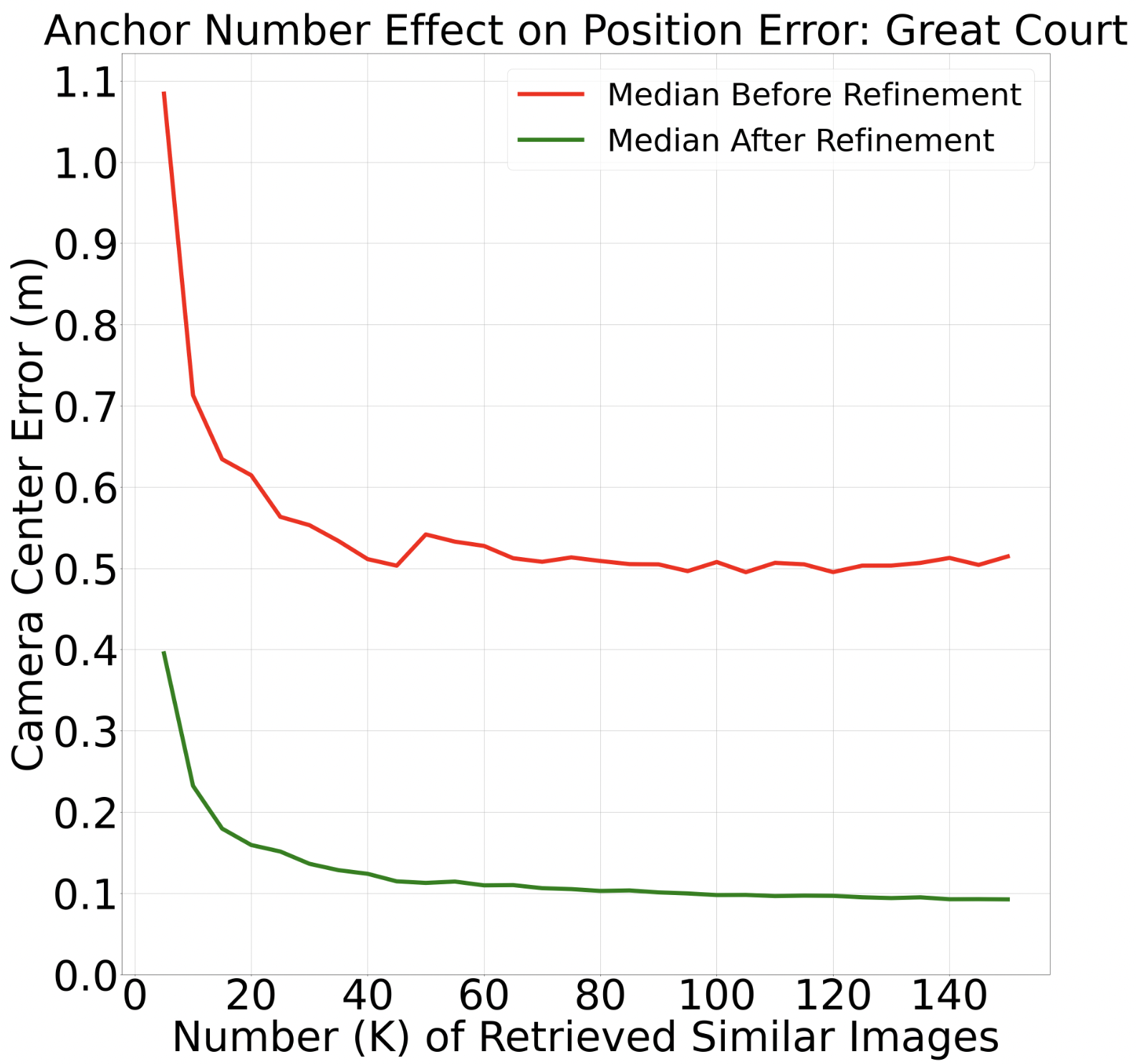}}
    (b) \fbox{\includegraphics[width=.4\linewidth]{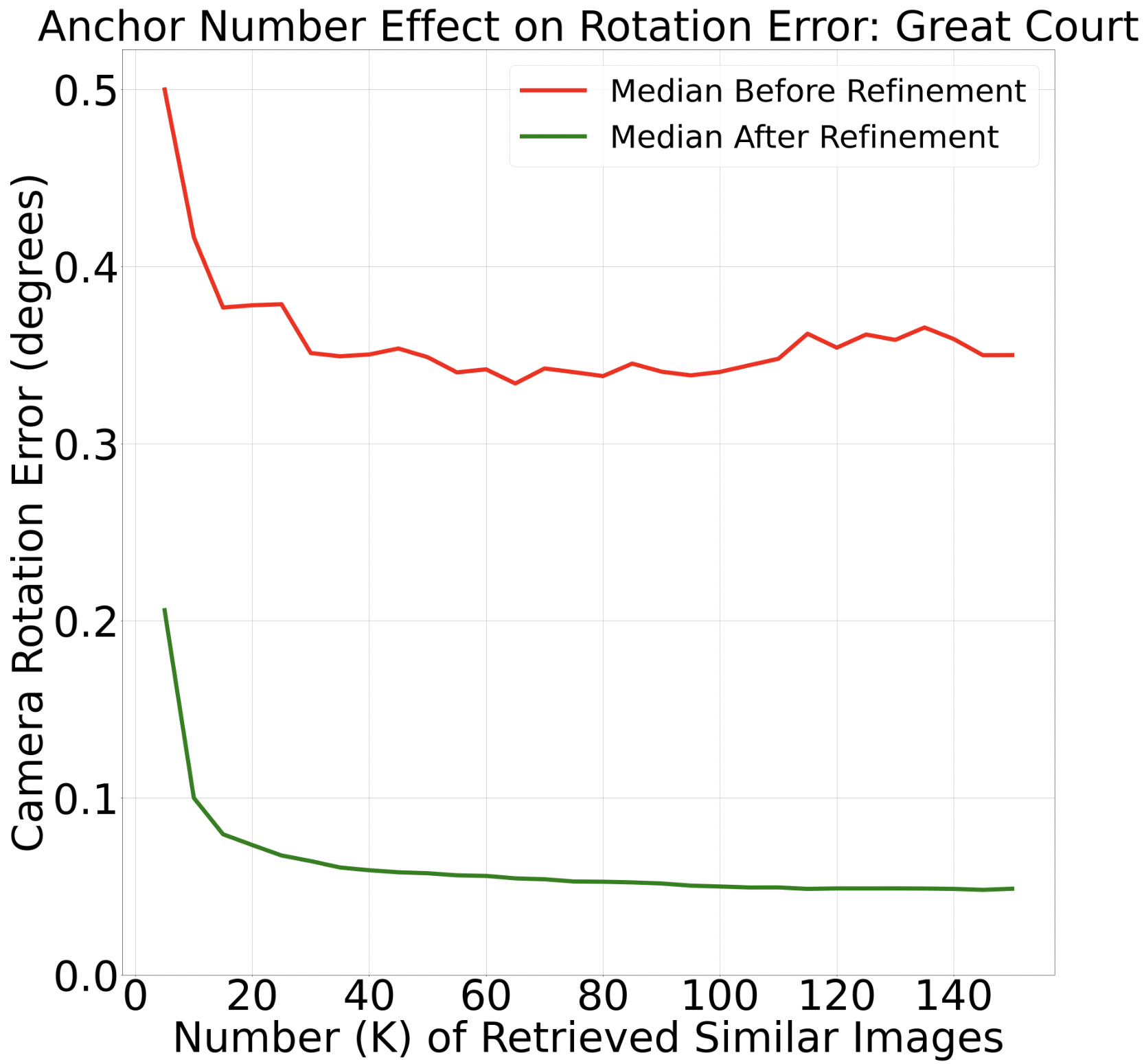}}}
    \caption{This study on the Great Court scene of Cambridge Landmarks shows that error drops significantly with increasing K up to around 50 images, but continues to decrease with increasing K for the range studied. }
    \label{fig:ksweep}
    \vspace{-0.2cm}
\end{figure}

\noindent \textbf{How Many Images Are Enough:} The typical perception casts doubt on the utility of additional images, beyond the first few, in improving localization: How could a new image contribute beyond the first ten or even twenty? It may be counter-intuitive, but Figure \ref{fig:ksweep} demonstrates that the second dozen of anchor images contribute significantly beyond the first dozen of anchor images. The incremental improvement drops in significance, but even the contribution of the second 50 images is not insignificant.

\noindent \textbf{Localization on Real Datasets:} Perhaps the most important evaluation is to measure the localization accuracy when compared to competing methods, Tables \ref{tab:results_7} and \ref{tab:results_aachen_robot}. Observe that in the 7-Scenes our approach is the top performer in five categories, nearly tied in one, and trailing in one. In the Cambridge Landmarks dataset, our approach almost sweeps the top performance in all scenes except one, where it is close to the top. The performance on the Aachen dataset is close to the best, performing as the second-best algorithm. The performance on the RobotCar dataset is at par for the day images, but has poor performance for the night images. As mentioned earlier, our approach requires numerous neighboring images to be effective, and the RobotCar dataset lacks in this aspect. 

\noindent \textbf{Timing and Storage:} 
The computational efficiency of our approach, as summarized in Figure~\ref{fig:main}, can be examined for the different stages. The image retrieval stage is fairly standard and quick, similar to the state of the art HLOC+SG~\cite{hloc,superglue}. The bottleneck in our scheme is the relative pose estimation which requires matching local features. Overall, our timing improves 5\% over HLOC+SG, Table~\ref{tab:timing}. In terms of memory footprint, our approach improves on HLOC+SG by~10\%. These measurements are for research code which has not yet been optimized. 

\setlength\tabcolsep{1pt} 
\begin{table}[h]
{\scriptsize
    \centering
    \begin{tabular}{|c|c|c|c|c|c|c|c|}
    \hline
    Pipeline & NV & SP & Retrieval & Matching (SG) & Localization & Total & Storage (GB) \\
    \hline
    HLOC + SG & 
    3 & 
    14 & 
    7 & 
    641 & 
    1333 &
    1998 & 
    8.544\\
    \hline
    \textbf{Ours} & 
    3 & 
    14 & 
    7 & 
    1538 & 
    361 &
    1903 & 
    7.431 \\
    \hline
    \end{tabular}
    \caption{\textbf{Timing [ms] and Storage [GB]:} The timing for the different stages of our pipeline as compared to the state-of-the-art HLoc + SG pipeline~\cite{hloc,superglue} (50 references images for retrieval). The stages measured are NetVLAD descriptor generation (NV), SuperPoint local feature finding (SP), image retrieval, local feature matching, and finally localization. The table shows the time required to localize one Aachen-day query. In terms of storage, both ours and HLoc require storing local and global descriptors for all reference images, but HLoc also stores an sfm reconstruction that requires an additional 1.131 GB.}
    \label{tab:timing}
}
\end{table}

\section{Conclusion}
This paper has proposed two key ideas: (\emph{i}) a camera center averaging scheme that relies on a coupling of orientation and localization, and (\emph{ii}) a latent 3D scheme for pose estimation directly from features once an accurate initial pose is available. Both ideas contribute to improved localization accuracy while also improving on computational efficiency, provided there is a sufficient number of images viewing the scene.


{
    \small
    \bibliographystyle{ieee_fullname}
    \bibliography{main}
}
\end{document}